\documentclass[11pt]{article} 
\usepackage{listings}
\usepackage[utf8]{inputenc} 
\usepackage{amsfonts}
\usepackage{amsmath}

\usepackage{geometry} 
\usepackage{graphicx} 
\usepackage{natbib}

\usepackage{booktabs} 
\usepackage{array} 
\usepackage{paralist} 
\usepackage{verbatim} 
\usepackage{subfig} 

\usepackage{fancyhdr} 
\newcommand{\trace}{\mathop{\mathrm{Tr}}}

\newcommand{\R}{\mathbb{R}}
\newcommand{\N}{\mathbb{N}}

\newcommand{\CC}{\mathcal{C}^{\mathrm{clean}}}
\newcommand{\CL}{\mathcal{L}^{\mathrm{clean}}}
\usepackage{sectsty}
\allsectionsfont{\sffamily\mdseries\upshape} 

\usepackage[nottoc,notlof,notlot]{tocbibind} 
\usepackage[titles,subfigure]{tocloft} 

\title{Adding noise to the input of a model trained with a regularized objective }
\author{Salah Rifai, Xavier Glorot, Yoshua Bengio and Pascal Vincent}

\begin{document}

\maketitle
\begin{center} Dept. IRO, Universit\'e de Montr\'eal. Montr\'eal (QC), H3C 3J7, Canada \end{center}
\begin{center}		Technical report 1359 \end{center}
\begin{abstract}
\noindent Regularization is a well studied problem in the context of neural networks. It is usually
used to improve the generalization performance when the number of input samples is relatively
small or heavily contaminated with noise. The regularization of a parametric model can be
achieved in different manners some of which are early stopping~\citep{Morgan+Bourlard90b}, weight decay, output
smoothing that are used to avoid overfitting during the training of the considered model.
From a Bayesian point of view, many regularization techniques correspond to imposing certain
prior distributions on model parameters~\citep{Krogh91}. Using Bishop's approximation~\citep{bishop95training} of the objective 
function when a restricted type of noise is added to the
input of a parametric function, we derive the higher order terms of the Taylor expansion and analyze the coefficients
of the regularization terms induced by the noisy input. In particular we study the effect of penalizing the Hessian
of the mapping function with respect to the input in terms of generalization performance. We also show how we can
control independently this coefficient by explicitly penalizing the Jacobian of the mapping function on corrupted
inputs.

\end{abstract}
\section{Introduction}
\label{intro}

Regularization is a well studied problem in the context of neural networks. It is usually
used to improve the generalization performance when the number of input samples is relatively
small or heavily contaminated with noise. The regularization of a parametric model can be
achieved in different manners some of which are early stopping~\citep{Morgan+Bourlard90b}, weight decay or output
smoothing, and are used to avoid overfitting during the training of the considered model.
From a Bayesian point of view, many regularization techniques correspond to imposing certain
prior distributions on model parameters~\citep{Krogh91}.

In this paper we propose a novel approach to achieve regularization that combines noise in the
input and explicit output smoothing by regularizing the L2-norm of the Jacobian's mapping function 
with respect to the input. \citet{bishop95training} has proved that the two approaches are essentially
equivalent under some assumptions using a Taylor approximation up to the second order of the
noisy objective function. Using his theoretical analysis, we derive the approximation of our
cost function in the weak noise limit and show the advantage of our technique from the theoretical
and empirical point of view. In particular, we show that we achieve a better smoothing of the
output of the considered model with a little computational overhead.

\section{Definitions}
For the ease of readability, most of our analysis involves only vectors and matrices except for
section~\ref{sec:higherterm} for which it was not possible to avoid using tensors objects. Also our analysis assumes
that the model's output is scalar which will prevent the use of tensors for the low order terms
of the Taylor expansion.
We will use the following notations:

\begin{itemize}
\item $\left<.,.\right>$ : inner product,
\item $\otimes$ : tensor product,
\item $J_f(x)$, $H_f(x)$, $T^{(n)}_f(x)$ : respectively the Jacobian, Hessian and $n$-th order derivative of $f$ with respect to vector $x$.
\end{itemize}

We consider the following set of points:
\[
\mathcal{D}_n=\left\{z_i=(x_i,y_i) \in (\mathbb{R}^d,\mathbb{R}) |  \forall i \in \left[\kern-0.15em\left[1;n\right]\kern-0.15em\right]\right\}
\]
where the $(x_i,y_i)$ are the (input,target) of an arbitrary dataset.
In the paper we will consider a particular family of parametric models
\[
\mathcal{F}= \left\{ F_{\theta} \in \mathrm{C}^\infty\left(\R^d\right)\mid \theta \in \R^p \mid p \in \N \right\}
\]
and $F_{\theta}(\R^d) \subseteq \R$.
We define the expected cost of the true distribution $p(z)$ of our data points as being:
\begin{equation}
\label{eq:expcost}
\mathcal{C}(\theta) = \int \mathcal{L}(z,\theta)p(z)dz
\end{equation}
The expected empirical cost when using the data without noise can be expressed as:
\begin{equation}
\label{eq:cleancost}
\mathcal{C}^{\mathrm{clean}}(\theta) = \int \mathcal{L}(z,\theta)\delta(z_i-z)dz = \frac{1}{n}\sum_i^n \mathcal{L}(z_i,\theta)
\end{equation}
where $\delta$ is the Dirac function.
When adding noise to the input, we will consider $p(z)$ as being a parzen density estimator:
\begin{equation}
\label{eq:parzen}
\displaystyle p(z) = \frac{1}{n} \sum_i^n\mathcal{\psi}(z_i-z)
\end{equation}
with the kernels $\psi$ centered on the points of our dataset.
In the rest of this paper we will consider kernels for which the following assumptions hold:
\begin{enumerate}[(a)]
\item every kernel has zero mean,
\item different components of a kernel are independent.
\end{enumerate}
Note that the normal and uniform distribution have these properties.
Using (a), we can write~\citep{An96}:
\begin{equation}
\label{eq:kernzero}
\forall i, \int \varepsilon_i \psi(\varepsilon)d\varepsilon =0
\end{equation}
and using (b) :
\begin{equation}
\label{eq:kernind}
\forall (i,j), \int \varepsilon_i \varepsilon_j \psi(\varepsilon) d\varepsilon = \sigma^2 \delta_{ij}
\end{equation}
where $\sigma$ is the variance of the distribution $\phi$, and $\varepsilon = (\varepsilon_1,\ldots,\varepsilon_d)$.
In our analysis we will restrict ourselves to gaussian kernels:
\begin{equation}
\label{eq:gauss}
\psi(z_i-z) = \mathcal{N}_{z_i,\sigma^2}(z)
\end{equation}

Substituting (\ref{eq:gauss}) in (\ref{eq:parzen}) we can write the objective function with noisy input as being:
\begin{equation}
\label{eq:noisycost}
\mathcal{C}^{\mathrm{noisy}}(\theta) = \frac{1}{n}\sum_i^n \int \mathcal{L}(z,\theta)\mathcal{N}_{z_i,\sigma^2}(z)dz
\end{equation}

\section{Penalty term induced by noisy input}

\citet{bishop95training} already showed that tuning the parameters of a model with corrupted inputs is asymptotically
equivalent to minimizing the true error function when simultaneously decreasing the level of corruption
to zero as the number of corrupted inputs tends to infinity. He also showed that adding noise in the input
is equivalent to minimize a different objective function that includes one or more penalty terms, and he uses
a Taylor expansion to derive an analytic approximation of the noise induced penalty.
Using the above assumption we can write:
\begin{equation}
\label{eq:penalty}
\mathcal{C}^{\mathrm{noisy}}(\theta) = \mathcal{C}^{\mathrm{clean}}(\theta) + \phi(\theta)
\end{equation}
where $\phi$ is the penalty term.
Substituting  (\ref{eq:noisycost}) and (\ref{eq:cleancost}) in (\ref{eq:penalty}) we express the
penalty term as being:
\begin{equation}
\label{eq:difference}
\phi(\theta)= \frac{1}{n}\sum_i^n \left[ \int \mathcal{L}(z,\theta)\mathcal{N}_{z_i,\sigma^2}(z)dz - \mathcal{L}(z_i,\theta) \right]
\end{equation}
We define the noise vector as being $\varepsilon = z - z_i$ and omitting $\theta$ for simplicity
we can write $\forall i$, the term inside the sum of (\ref{eq:difference}):
\begin{equation}
\label{eq:simplediff}
D =  \int \mathcal{L}(z_i + \varepsilon)\mathcal{N}_{0,\sigma^2}(\varepsilon)d\varepsilon  -\mathcal{L}(z_i)
\end{equation}
Now that we have identified the term to approximate, let's write the Taylor approximation of our loss
 function when our sample is shifted by a noise vector $\varepsilon$:
\begin{equation}
\label{eq:taylor}
\mathcal{L}(z + \varepsilon) =  \mathcal{L}(z)  + \langle J_{\mathcal{L}}(z), \varepsilon \rangle +\frac{1}{2} \varepsilon^T . H_{\mathcal{L}}(z) . \varepsilon + o(\varepsilon^2)\
\end{equation}
To match equation (\ref{eq:simplediff}) we multiply with (\ref{eq:gauss}) and integrate with respect to $\varepsilon$ both sides of (\ref{eq:taylor}):
\begin{equation}
\begin{array}{c}
\displaystyle \int \mathcal{L}(z + \varepsilon)\mathcal{N}_{0,\sigma^2}(\varepsilon)d\varepsilon = \label{eq:taylor1}\\
\cr\displaystyle \int \left[  \mathcal{L}(z) + \langle J_{\mathcal{L}}(z), \varepsilon \rangle +
\frac{1}{2} \varepsilon^T . H_{\mathcal{L}}(z) . \varepsilon + o(\varepsilon)\right]
\mathcal{N}_{0,\sigma^2}(\varepsilon)d\varepsilon
\end{array}
\end{equation}
Equation (\ref{eq:kernzero}) implies that all odd-moments of the approximation are null and in conjunction
with (\ref{eq:kernind}) we can now simplify (\ref{eq:taylor1}) into:
\begin{equation}
\begin{array}{l}
\displaystyle \int \mathcal{L}(z + \varepsilon)\mathcal{N}_{0,\sigma^2}(\varepsilon)d\varepsilon \cr \\
\cr = \mathcal{L}(z)\underbrace{\int \mathcal{N}_{0,\sigma^2}(\varepsilon)d\varepsilon}_\textrm{1} +
\underbrace{\int \langle J_{\mathcal{L}}(z), \varepsilon \rangle \mathcal{N}_{0\,\sigma^2}(\varepsilon)d\varepsilon}_\textrm{0} \\
\cr \displaystyle +  \frac{1}{2}\int  \varepsilon^T . H_{\mathcal{L}}(z) . \varepsilon  \mathcal{N}_{0,\sigma^2}(\varepsilon)d\varepsilon + R \label{eq:taylor2}
\end{array}
\end{equation}
and with some algebra we can finally write:
\begin{equation}
\label{eq:taylorend}
\int \mathcal{L}(z + \varepsilon)\mathcal{N}_{0,\sigma^2}(\varepsilon)d\varepsilon - \mathcal{L}(z)  \approx \frac{\sigma^2}{2} \trace(H_{\mathcal{L}}(z))
\end{equation}
by substituting $z=z_i$ and summing over all the elements of $\mathcal{D}_n$:
\begin{equation}
\label{eq:phi}
\phi(\theta) \approx \frac{\sigma^2}{2n}\sum_i^n \trace(H_{\mathcal{L}}(z_i))
\end{equation}
Hence training with corrupted inputs is approximately equivalent to minimize the following objective:
\begin{equation}
\label{eq:noiseapprox}
  \addtolength{\fboxsep}{5pt}
   \boxed{
   \begin{gathered}
\mathcal{C}^{\mathrm{noise}}(\theta) \approx  \mathcal{C}^{\mathrm{clean}}(\theta) + \lambda \trace(H_{\mathcal{\bar{C}}}(\theta))
   \end{gathered}
}
\end{equation}

This relation holds for any objective $\mathcal{C}$ with the above two assumptions.
\newline\newline
All the above results are already well established in the literature of noise injection~\citep{bishop95training,An96}. Our contribution is to show that, for the second order Taylor expansion, adding noise to the input on a well chosen regularized objective is equivalent to add a $L_2$-norm on the Hessian of the output function of the considered model respectively to its input, which is not the case when adding noise to a non-regularized objective.

In the following sections we will consider the MSE as the objective function to tune the parameters of our model but this choice does not affect the generality of our analysis.
\section{Non regularized objective}
We define the following error function:
\[ \CL(z_i,\theta) = \parallel F(x_i,\theta) - y_i \parallel^2 \]
and the Hessian of the cost function being the average over the Hessian of all individual errors:
\[  H_{\CC}(\theta) = \frac{1}{n}\sum_i^n   H_{\CL}(z_i,\theta) \]
Assuming without loss of generality that $F$ is a scalar function\footnote{in the case of multiple regression such as a reconstruction function, only the orders of the tensors involved in the approximation of $\mathcal{L}$ will change.}, and that the noise is added to the input $x$, we will only consider the Hessian of the loss function with respect to $x$ :
\[ H_{\CL}(x) = \frac{\partial}{\partial x} \left[\frac{\partial}{\partial x} \left( \parallel F(x,\theta) - y \parallel^2 \right)\right]  \]
\[ =2   \frac{\partial}{\partial x} \left[ \frac{\partial F\left(x,\theta\right)}{\partial x}  \left(  F(x,\theta) - y \right)\right] \]
\[ =2 \left[  \frac{\partial^2 F\left(x,\theta\right) }{(\partial x)^2} \left(  F(x,\theta) - y \right) + \left<  \frac{\partial F\left(x,\theta\right)}{\partial x} \right>^2 \right]  \]
\[
H_{\CL}(x) = 2  H_{F}(x)   \left(  F(x,\theta) - y \right) + 2\left<J_F(x)^T, J_F(x)\right>
\]
A standard result of linear algebra is the relation between the Frobenius norm and the trace operator:
\[  \trace\left(\left<A^T,A\right>\right)  = \left|\left| A\right|\right|_F^2 \]
By taking the trace of the above results, we get:
\[ 
\trace\left(H_{\CL}(x)\right) = 2  \left(  F(x,\theta) - y \right) \trace\left(H_{F}(x)\right)   + 2\left|\left|  J_F(x) \right|\right|_F^2 
\]
and plugging this in (\ref{eq:noiseapprox}) gives us the following second order approximation of the noisy objective:
\begin{equation}
\label{eq:fullnoiseapprox}
\
  \addtolength{\fboxsep}{5pt}
   \boxed{
   \begin{gathered}
\mathcal{L}^{\mathrm{noise}}(z_i,\theta) \approx  \parallel F(x_i,\theta) - y_i \parallel^2+ 2\lambda \left(  \left(  F(x,\theta) - y \right) \trace\left(H_{F}(x)\right)   + \left|\left|  J_F(x) \right|\right|_F^2 \right)
   \end{gathered}
   }
\end{equation}
We obtain a $L_2$-norm on the gradient of the mapping function $F$ added to our error function whereas the Hessian term is not constrained to be positive and it is not sure that its terms are going to cancel-out. E.g. if prediction overshoots or undershoots on average, then penalty may encourage very large Hessian trace inducing high curvature which would potentially harm a stochastic gradient descent and converge to a poor local minimum.

\section{Our regularized objective}
As the results of the previous section suggest, adding noise to the input of the objective function yields an undesirable term that might interfere with the goal of smoothing the output of our function. We propose here to overcome this difficulty by adding noise only to the input of the objective function's Jacobian $J_F$, doing so will avoid the unpredictable effect of additional unwanted terms. 
We define the error function we previously used to which we add a regularization term that is the $L_2$-norm of the Jacobian of $F$ with respect to $x$ : 
\[ \mathcal{L}^{\mathrm{reg+noise}}(z_i,\theta) = \parallel F(x_i,\theta) - y_i \parallel^2 + \lambda\left|\left| J_F(\tilde{x}_i)  \right|\right|^2_F \]
We now calculate the Hessian of $\tilde{\mathcal{L}}$ by omitting the first term since the noise is added only to the input of the regularization term, and using the approximation derived in  (\ref{eq:noiseapprox}):
\[\left|\left|  \frac{\partial F\left(\tilde{x}_i,\theta\right)}{\partial x}  \right|\right|^2_F \approx 
\left|\left|  \frac{\partial F\left(x_i,\theta\right)}{\partial x}  \right|\right|^2_F  + 
\varepsilon^2 \trace\left( \displaystyle H_{\left|\left|  \frac{\partial F}{\partial x}  \right|\right|^2} (x)\right)
  \]
We can now calculate the Hessian of the regularization term as being:

\[
H_{\left|\left|  \frac{\partial F}{\partial x}  \right|\right|^2} (x)  =  2 \frac{\partial}{\partial x} \left[  \frac{\partial^2 F\left(x,\theta\right) }{(\partial x)^2} \frac{\partial F\left(x,\theta\right)}{\partial x} \right] \]
\[ = 2\left[  \frac{\partial^3 F\left(x,\theta\right) }{(\partial x)^3} \frac{\partial F\left(x,\theta\right)}{\partial x} +  \left<\frac{\partial^2 F\left(x,\theta\right) }{(\partial x)^2}\right>^2 \right] \]
\[
 H_{\left|\left|  \frac{\partial F}{\partial x}  \right|\right|^2}(x) = 2\frac{\partial^3 F\left(x,\theta\right) }{(\partial x)^3} J_F(x) +  2\left<H_F(x)^T,H_F(x)\right>
\]

\begin{equation}
\trace\left(H_{\left|\left|  \frac{\partial F}{\partial x}  \right|\right|^2} (x) \right) = 2 \trace\left(\frac{\partial^3 F\left(x,\theta\right) }{(\partial x)^3} J_F(x)\right) + 2\left|\left|  H_F(x) \right|\right|_F^2
\end{equation}
which gives us the approximation of our regularized objective:
\begin{equation}
\label{eq:fullregapprox}
\
  \addtolength{\fboxsep}{5pt}
   \boxed{
   \begin{gathered}
\mathcal{L}^{\mathrm{reg+noise}}(z_i,\theta) = \parallel F(x_i,\theta) - y_i \parallel^2 +
 \lambda\left|\left| J_F(x)  \right|\right|^2_F +
 2\lambda\sigma^2\left|\left|  H_F(x) \right|\right|_F^2 + R
  \end{gathered}
   }
\end{equation}

We have shown that adding noise to a well chosen regularized objective clearly penalize the 
$L_2$-norm on the Hessian of the considered model $F$(without ever calculating it) using a 
second order Taylor approximation of the noisy objective under two necessary assumptions on 
the noise distribution. In statistics regularizing the norm of the derivatives of the model
to be tuned is often referred as roughness penalty~\citep{Green:1993} and is used in the context
of cubic splines~\citep{Boor98}.
\section{Higher order terms of the Taylor expansion}
\label{sec:higherterm}
In this section we are interested in the higher order terms of the cost approximation, we find it convenient to use the following formalism:
\newline
if $T_{\mathcal{L}}^n(z,\varepsilon)$ denotes the n-th order derivative of $\mathcal{L}$ with respect to $z$, then:
\[ T_{\mathcal{L}}^n(z,\varepsilon) = \frac{1}{n!} \sum_{i_1,\ldots,i_n}  \varepsilon_{i_1},\ldots,\varepsilon_{i_n}T_{i_1,\ldots,i_n}^n (z) \] 
where $T^n$ is a tensor of order $n$ and
\[ T_{i_1,\ldots,i_n}^n(z) = \frac{\partial^n \mathcal{L} (z)}{\partial z_{i_1},...,\partial z_{i_n}} \] 
using this formalism we can write the fourth order derivative as being:
\[ T^4_{\mathcal{L}}(z,\varepsilon) = \frac{1}{24} \sum_{i,j,k,l} \varepsilon_i\varepsilon_j\varepsilon_k \varepsilon_l T^4_{ijkl}(z) \]
Using the two assumptions made on the noise distribution, we know that the third order derivative of the approximation is zero. As for the fourth order derivative, using the second assumption of the noise distribution we know that only the terms that are on the diagonal of the $T^4$ will be non-zero, we can then write:

\begin{equation*}
  \addtolength{\fboxsep}{5pt}
   \boxed{
   \begin{gathered}
\int  T^4_{\mathcal{L}}(z,\varepsilon) \mathcal{N}_{0,\sigma^2}(\varepsilon)d\varepsilon =  \frac{\sigma^4}{4!} \sum_{i}T^4_{iiii}(z) 
   \end{gathered}
   }
\end{equation*}

Using the above result we can approximate our cost function in the noisy input setting more finely, for this purpose we will use the results obtained above for the Hessian and differentiate them again twice with respect to $x$.
\[ T^4_{\mathcal{L}}(x) = \frac{\partial^2\left(H_{\mathcal{L}}\left(x\right)\right) }{(\partial x)^2} \]
\[ = \frac{\partial^2}{(\partial x)^2}\left[ 2  H_{F}(x)   \left(  F(x,\theta) - y \right) + 2\left<J_F(x)^T, J_F(x)\right>\right] \]
\[ = 2\frac{\partial}{\partial x}\left[T^3_{F}(x)  \left(  F(x,\theta) - y \right) + 3 H_{F}(x)J_{F}(x) \right] \]
\[ = 2\left[ T^4_{F}(x)  \left(  F(x,\theta) - y \right) + 4T^3_{F}(x)J_{F}(x) + 3\left<H_F(x)^T, H_F(x)\right>\right] \]
hence,
\begin{equation*}
  \addtolength{\fboxsep}{5pt}
   \boxed{
   \begin{gathered}
 \trace\left(T^4_{\mathcal{L}}\left(z\right)\right) = 6\left|\left|  H_F(x) \right|\right|_F^2 + \trace(R)
   \end{gathered}
   }
\end{equation*}

where $R = 2T^4_{F}(x)  \left(  F(x,\theta) - y \right) + 8T^3_{F}(x)J_{F}(x)$. 
\section{Comparison}
\subsection{Noise added to the input of the objective function}
Now that we have a higher order approximation in the case where noise is added to the input of the function, we can compare the magnitude of the coefficients that penalize the Hessian $H_F$, note that in this case the Hessian term appears in the fourth order of the Taylor expansion of the cost function, whereas we need only a second order approximation in the case where we add a regularization term evaluated on a corrupted input.
We can write the approximation of the cost function without regularization at the fourth order as being:

\[\mathcal{L}^{\mathrm{noise}}(z) \approx  \mathcal{L}^{\mathrm{clean}}(z) +  \frac{\sigma^2}{2!}  \trace(H_{\mathcal{L}^{\mathrm{clean}}}(z)) +  \frac{\sigma^4}{4!}  \trace\left(T^4_{\mathcal{L}^{\mathrm{clean}}}\left(z\right)\right) 
\]
\begin{equation}
  \addtolength{\fboxsep}{5pt}
   \boxed{
   \begin{gathered}
\mathcal{L}^{\mathrm{noise}}(z) =   \parallel F(x_i,\theta) - y_i \parallel^2 + \overline{\sigma^2 \left|\left|  J_F(x) \right|\right|_F^2} +  
\frac{\sigma^4}{4} \overline{\overline{\left|\left|  H_F(x) \right|\right|_F^2}} + \overline{\overline{R}}
   \end{gathered}
   }
\end{equation}
where the number $i$ of overline denotes the terms induced by the noise obtained at the $i$-th order of the Taylor expansion.
\subsection{Noise added to the input of the Jacobian of the objective function}
In this case we just need to approximate the cost function up to the second order of the Taylor expansion:
\begin{equation}
\label{eq:fullregapproxfinal}
\
  \addtolength{\fboxsep}{5pt}
   \boxed{
   \begin{gathered}
\mathcal{L}^{\mathrm{reg+noise}}(z_i,\theta) = \parallel F(x_i,\theta) - y_i \parallel^2 +
 \lambda\left|\left| J_F(x)  \right|\right|^2_F +
 \overline{2\lambda\sigma^2\left|\left|  H_F(x) \right|\right|_F^2} + \overline{R}
  \end{gathered}
   }
\end{equation}

\section{Experimental results}
We have tried several experiments in order to benchmark the effect of regularization and noise combined, for this task we used the well known MNIST~\citep{LeCun+98}, MNIST binarised and the USPS database. Surprisingly, we were able to achieve results close to those obtained with unsupervised pretraining~\citep{Erhan2010}. 
MNIST is composed of $70k$ handwritten digits
represented by a vector of pixels. It is divided in $50k$ for the training set, $10k$ for each of the validation and test set, the range of the features were rescaled to be
within  $\left[0,1\right]$. MNIST-Binary is divided exactly the same way as MNIST, the only difference is that the intensity of the pixels superior to $\frac{255}{2}$ where
set to $1$ and the others to $0$. USPS dataset consists of a training set with 7291 images and a test set with 2007 images, the validation set was 
formed using the last 2000 images of the training set. The model $F(x)$, with parameters $\theta=\{W^{(1)},\ldots,W^{(n)},b^{(1)},\ldots,b^{(n)}\}$, we considered to solve the classification task was a standard neural network with one or more hidden layers, and a hyperbolic tangent non-linearity in between the layers. For example, with one hidden layer we have:
\[ F(x) = \sigma(W^{(2)}\tanh(W^{(1)}.x+b^{(1)})+ b^{(2)}) \]

where $\sigma(.)$ is the logistic sigmoid function. In this setting we can write the Frobenius norm of the Jacobian of $F$ with respect to $x$ as being:
\[ \left|\left|J_F(x)\right|\right|^2 =  \sum_{i,j} J_{ij}^2 \]

and with some calculus we get:
\[ J_{ij} = F_i(x)(1-F_i(x))\sum_l W^{(2)}_{il} W^{(1)}_{lj} \left(1-\tanh^2\left(\sum_m W^{(1)}_{lm}x_m + b^{(1)}_l\right)\right) \]

where $F_i(x)$ is $i$-th output of the network.
For the results in table (\ref{table:res}), we used a number of hidden units ranging from $400$ to $1000$ per layer, 
the best results were obtained with two hidden layers on MNIST, and one hidden layer on MNIST-BINARY and USPS. 
The parameters of the model were learned through stochastic gradient descent with a learning rate ranging from $0.1$ to $0.001$.
We also investigated the use of {\em Rectifying} units (i.e. $\max(0,x)$)~\citep{Hinton2010,Glorotwkp:2010} as non-linearity in the hidden layer, surprisingly they seemed to benefit less from the added noise to the input
than from the regularization term alone, they achieved a test classification performance of $4.8\%$ on the USPS dataset equaling the best performance of the hyperbolic
tangent activation with both regularization and added noise to the input. The best results where obtained with a gaussian isotropic noise with a standard deviation
of $0.1$ around training samples.

\begin{figure*}[ht]
  \begin{center}
    \resizebox{\textwidth}{!}{\includegraphics{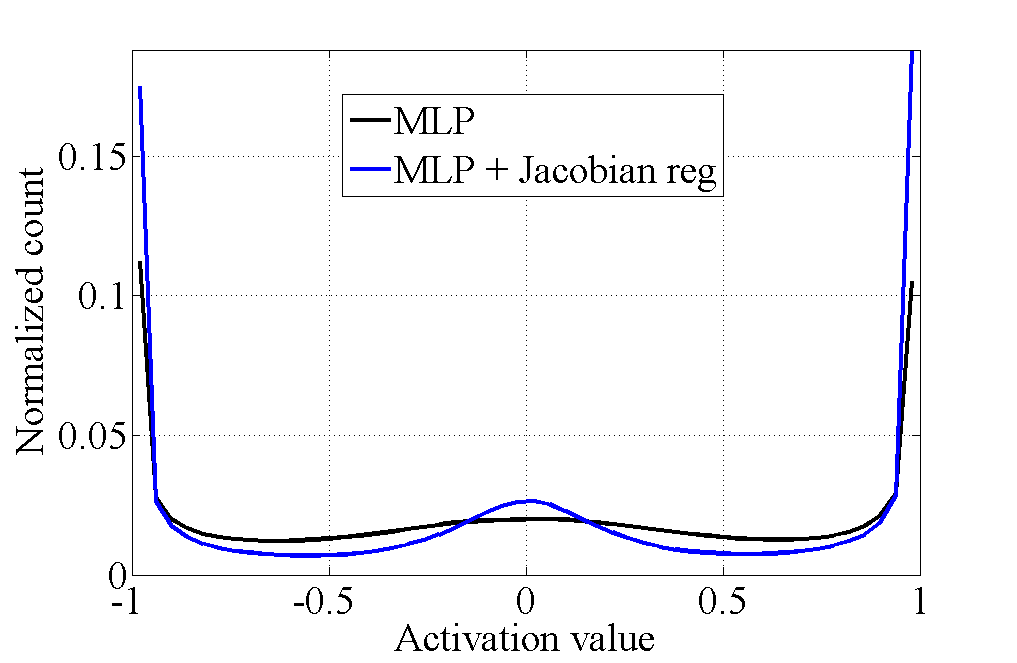}}
    \caption{Normalized histograms of the activation values
    on the MNIST test set for the best MLPs with and without
    Jacobian regularization. The activations of the regularized 
    model are more densely distributed at the saturation and
    linear regime.}
    \label{fig:act}
    \end{center}
\end{figure*}

Figure~\ref{fig:act} shows the histograms of activation values on the MNIST test set of our best MLPs with and without Jacobian regularization.
The activations of the regularized model are more densely distributed at the saturation and linear regime.

\section{Discussion}
\subsection{Constraining the solution space}
\begin{figure*}[ht]
  \begin{center}
    \resizebox{\textwidth}{!}{\includegraphics{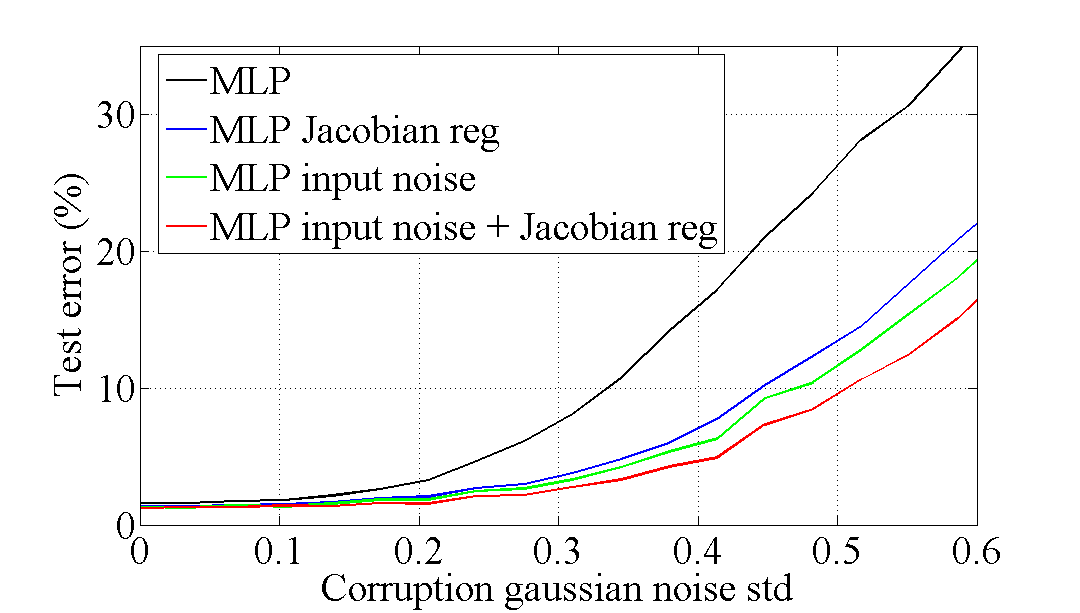}}
    \caption{Robustness of the generalization error with respect
    to a gaussian corruption noise added to the input, 
    the model trained with the combination of
    input noise and Jacobian regularization is more robust.}
    \label{fig:noise}
    \end{center}
\end{figure*}

When optimizing a non-convex function with an infinite amount of local minimum, it is not clear which of them yields a reasonable generalization performance,
the concept of overfitting clearly illustrate this point. The proposed regularizer tries to avoid this scheme by flattening the mapping function
over the training examples inducing a local invariance in the mapping space to infinitesimal variance in the input space. Figure \ref{fig:noise} shows that when
the input is corrupted, the models learned with the regularization term are more robust to noise on the input. Geometrically, the added regularization imposes
to the model to be a {\em Lipschitz} function or a contracting map around the training examples imposing the constraint $F(x+\varepsilon) \approx F(x)$.
\subsection{Smoothing away from the training points}

Penalizing only the Jacobian of the model $F$ with respect to the input $x$ gives only a guarantee of flatness for an infinitesimal move away of the training example.
To illustrate this point one can imagine that on the tip of a dirac function the norm of the jacobian is null and infinite just around it. Although this situation is
not possible in the context of neural networks because of their smooth activation function. Given enough capacity we could converge to this solution if we do not add additional
constraints. One of them would be to adjust the locality of the flatness as a hyper-parameter of the model. It requires to compute the higher order terms of 
the mapping function in order to regularize the magnitude of their norms. As it was discussed before, it is computationally expansive to explicitly calculate the norm 
of the high order derivatives because their number of components increases exponentially, instead our approach proposes an approximation of the Hessian term that allows you 
to simultaneously control the magnitude of both the Jacobian and hessian norms independently.

\subsection{The other terms induced by the noise}
As we have seen in the equation \ref{eq:fullnoiseapprox}, the added noise does not yield only in a penalty on the norm of the successive derivatives of the mapping function and
it is somehow unclear how these terms behave during gradient descent since they are not constrained to be positive.
In a supervised setting, it is empirically feasible to drive those terms to zero because of the small dimensionality of the target points, 
whereas in a multi-dimensional regression task such as the reconstruction objective of an auto-encoder it is 
often impossible to achieve a ``near'' zero minimization of the cost with a first order optimization such as a stochastic gradient descent.
The reader should note that the approximation of the noisy cost is valid when the number of corrupted inputs tends to infinity, though in practise this is never the case.
It would be interesting to do an estimate of the difference between the terms induced by the noise and the real values of the term in function of the number of corrupted samples.

\section{Conclusion}
We have shown how to obtain a better generalization performance using a regularization term that adds a marginal computational overhead
compared to the traditional approach. Using a Taylor expansion of the cost function, we also showed that by adding noise to the input 
of the regularization term we are able to penalize with a greater magnitude the norm of the higher order derivatives of the model
avoiding the need to explicitly calculate them, which would be obviously computationally prohibitive. Initial results suggests that
different parametric models clearly benefit from this approach in terms of predicting out-of-sample points. It would be interesting to 
investigate how this regularization approach would behave when used with non-parametric models such as gaussian-mixtures.

\begin{table}[t]
\caption{Test error}
\label{table:res}
\begin{center}
\begin{tabular}{|l|c|c|c|}\hline
{\bf Model}    &{\bf MNIST}      &{\bf MNIST-BINARY}    &{\bf USPS}\\ \hline
MLP             &{1.82\%}        & {2.01\%}             & {5.74\%}\\ \hline
MLP + $L_2$     &{1.66\%}        & {2.20\%}             & {5.6\%}\\ \hline
MLP + Noise     &{\bf 1.33\%}    & {1.7\%}              & {4.85\%}\\ \hline
MLP + Jacob     &{\bf 1.31\%}    & {1.65\%}             & {4.85\%}\\ \hline
MLP + N + J     &{\bf 1.19\%}    & {\bf 1.51\%}         & {\bf 4.6\%}\\ \hline
\end{tabular}
\end{center}
\end{table}

{\small
\bibliography{strings,strings-short,strings-shorter,aigaion-shorter,ml,myrefs}
\bibliographystyle{natbib}
}

\end{document}